\documentclass{article}
\PassOptionsToPackage{numbers, compress}{natbib}

\usepackage[preprint]{neurips_2026}

\usepackage[utf8]{inputenc} 
\usepackage[T1]{fontenc}    
\usepackage{hyperref}       
\usepackage{url}            
\usepackage{booktabs}       
\usepackage{amsfonts}       
\usepackage{nicefrac}       
\usepackage{microtype}      
\usepackage{xcolor}         

\usepackage{graphicx}
\usepackage{subfigure}
\usepackage{booktabs} 

\usepackage{amsmath}
\usepackage{amssymb}
\usepackage{mathtools}
\usepackage{amsthm}
\usepackage{makecell}
\usepackage{color}
\usepackage{array}
\usepackage{caption}
\usepackage{graphicx}
\usepackage{subcaption}
\usepackage{tcolorbox}
\usepackage{multirow}
\usepackage{nccmath}
\usepackage{booktabs}
\usepackage{wrapfig}
\usepackage{enumitem}
\theoremstyle{plain}

\theoremstyle{definition}

\theoremstyle{remark}

\newtheorem*{remark*}{Remark}

\usepackage{bbm}
\usepackage{bm}
\usepackage[numbers]{natbib}
\title{Composable Crystals: Controllable Materials Discovery via Concept Learning}

\author{
Nian Liu$^{1}$ \quad \textbf{Yuwei Zeng}$^{1}$ \quad \textbf{Ryoji Kubo}$^{1}$ \quad Nikita Kazeev$^{1}$ \quad Stephen Gregory Dale$^{1}$\vspace{0.10cm}\\
 \quad \textbf{Artem Maevskiy}$^{1}$ \quad \textbf{Pengru Huang}$^{1}$ \quad \textbf{Thomas Laurent}$^{2}$ \vspace{0.10cm}\\
\textbf{Kostya S. Novoselov}$^{1}$ \quad \textbf{Xavier Bresson}$^{1}$
\vspace{0.15cm}\\
\texttt{\{nianliu, yuweizeng, ryojikubo\}@u.nus.edu, tlaurent@lmu.edu}\vspace{0.0cm}\\
\texttt{\{kna, sdale, maevskiy, pengru, kostya, xaviercs\}@nus.edu.sg }\vspace{0.15cm}\\
{\normalfont $^1$National University of Singapore \hspace{0.1cm} $^2$Loyola Marymount University \hspace{0.1cm}}\\
}

\begin{document}
\maketitle
\begin{abstract}
De novo crystal generation, a central task in materials discovery, aims to generate crystals that are simultaneously valid, stable, unique, and novel. Existing methods mainly rely on black-box stochastic sampling, providing limited control over how generated structures move beyond the observed distribution. In this paper, we introduce a concept-based compositional framework for crystal generation. We train a vector-quantized variational autoencoder to automatically discover a shared set of reusable crystal concepts, which serve as building blocks for guided generation. These learned concepts naturally exhibit interpretability from both local atomic environments and global symmetry patterns, and generalize to crystals from different distributions. By recombining such concepts, our framework enables controllable exploration of novel crystals beyond the training distribution, rather than relying solely on unconstrained random sampling. To further improve composition efficiency, we introduce a composition generator and iteratively refine it using high-quality samples generated by the model itself. The resulting concept compositions are then used to condition downstream crystal generation. Numerical experiments on MP-20 and Alex-MP-20 show that compositing concepts separately increase base model up to 53.2\% and 51.7\% on \texttt{V.S.U.N} metric, with particular gains in novelty. Our code is available at \url{https://github.com/liun-online/Compositional_Crystal_Generation}
\end{abstract}

\section{Introduction}
\label{intro}
Large models have become deeply integrated into everyday life. Users express their needs through textual prompts, and the models generate corresponding outputs. These prompts can often be understood as different compositions of a shared set of underlying concepts, such as dairy, veggie, meat, etc. Once a model learns to represent and interpret such concepts, it can generalize to unseen prompts by recombining them, even when the specific combinations were never observed during training~\cite{conceptmix}. In this work, we introduce this compositional generative paradigm into materials discovery. A primary objective in materials science is \textit{de novo generation} (DNG), which focuses on discovering novel, stable crystals without relying on predefined templates. Most prior studies~\cite{cdvae, diffcsp, adit} focus on unconditional generation. However, this paradigm provides limited control over the generation process: the initial states are sampled randomly, and the intermediate generation trajectory is often difficult to interpret. Such black-box stochasticity makes the discovery process difficult to control and systematically guide. In contrast, a compositional framework enables generation through the explicit recombination of concepts, offering a principled way to explore structures beyond the training distribution.

To realize compositional generation, the first question is how to define crystal concepts. We argue that ideal crystal concepts should satisfy four properties:
\begin{enumerate}
    \item \textit{Extractable}: Concepts should be readily extracted from any given crystal.
    \item \textit{Interpretable}: Some of these concepts produce discernible patterns that are meaningful and physically grounded.
    \item \textit{Generalizable}: The same concept set should remain applicable across crystals drawn from different distributions.
    \item \textit{Verifiable}: It should be easy to determine whether a generated crystal indeed contains the desired concepts.
\end{enumerate}
Fundamental physical properties, including band gap~\cite{mattergen} and formation energy~\cite{crystalformer}, constitute a natural choice for these underlying concepts. While these properties are interpretable and generalizable, obtaining their exact values typically requires expensive density functional theory (DFT) calculations~\cite{kresse1996vasp}, making both concept extraction and verification costly. In addition, meaningful composition of multiple physical properties often requires substantial domain expertise.

Once these concepts are defined, the next challenge is how to compose them into new candidates while filtering out invalid combinations. In text-to-image generation, \cite{conceptmix} randomly combined concepts and then validated the resulting compositions (i.e., textual prompts) using GPT-4o~\cite{gpt4o}. For crystals, however, even a subtle perturbation, such as changing a single atom type or slightly modifying atomic coordinates, may render a structure unstable. As a result, random concept composition is highly inefficient in this domain. Moreover, unlike language and vision, materials science currently lacks a foundation model that can automatically assess the validity of such compositions. Machine learning force fields, such as MatterSim~\cite{mattersim}, offer a path toward efficient stability evaluation, however, they do not solve the challenge of automated composition.

To address these challenges, we first train a vector-quantized variational autoencoder (VQ-VAE)~\cite{vqvae} on the MP-20 dataset~\cite{cdvae}, and treat the learned discrete latent codes as crystal concepts. These codes exhibit both local and global interpretability, and generalize to crystals from Alex-MP-20~\cite{mattergen}. We further verify that generated crystals are indeed composed of the learned concepts. To improve composition efficiency, we employ a denoising diffusion probabilistic model (DDPM)~\cite{ddpm} as a concept composition generator. We then validate the generated compositions by evaluating their V.S.U.N. (validity, stability, uniqueness, and novelty)~\cite{mattergen} with respect to the training set, and feed the qualified samples back to refine the composition generator. Finally, the generator produces multiple concept compositions to guide the downstream composition-based generative model in the DNG process. Our contributions are summarized as follows:
\begin{itemize}
    \item We introduce compositional generation into materials discovery, enabling controllable exploration beyond the training distribution.
    \item We propose crystal concepts that are simultaneously extractable, interpretable, generalizable, and verifiable.
    \item Conditioning on the proposed concept compositions improves the base generative model on both MP-20 and Alex-MP-20 by 53.2\% and 51.7\% on the \texttt{V.S.U.N} metric, where the main contribution is attributed to the gains in novelty.
\end{itemize}
\section{Preliminaries}
\label{pre}
\subsection{Crystal Representation}
A crystal $\mathbf{C}$ is defined by the periodic arrangement of its fundamental repeating unit, the unit cell, across the three-dimensional space. The unit cell can be fully described by Cartesian coordinates $\bm{X}\in\mathbb{R}^{N\times 3}$, atomic types $\bm{A}\in\mathbb{R}^{N}$, and a lattice matrix $\bm{L}\in\mathbb{R}^{3\times 3}$, where $N$ denotes the number of atoms in the unit cell. For $\bm{A}$, we consider the first 100 elements in the periodic table and encode them using one-hot vectors, denoted by $\text{one-hot}(\bm{A})\in\mathbb{R}^{N\times100}$. For $\bm{L}$, we follow prior work~\cite{mattergen} and decompose it via singular value decomposition (SVD), reparameterizing it as a symmetric matrix $\tilde{\bm{L}}$:
\begin{equation}
    \bm{L}=\bm{U}\tilde{\bm{L}},\quad \bm{U}=\bm{W}\bm{V}^{\top},\quad \tilde{\bm{L}}=\bm{V}\bm{\Sigma}\bm{V}^{\top},
\end{equation}
where $\bm{W}$ and $\bm{V}$ are the left and right singular vectors of $\bm{L}$, respectively, and $\bm{\Sigma}$ is the diagonal matrix of singular values. Here, $\bm{U}$ is a rotation matrix, and $\tilde{\bm{L}}$ is symmetric positive definite. We extract the upper-triangular entries of $\tilde{\bm{L}}$ and reshape them into a vector $\widehat{\bm{L}}=\mathrm{vec}(\mathrm{triu}(\tilde{\bm{L}}))\in\mathbb{R}^{6}$. The information associated with atom $i$ in $\mathbf{C}$ is summarized as an \textit{atom vector}:
\begin{equation}
\label{atom_rep}
    \bm{v}_i=[\bm{X}_i \, \| \, \text{one-hot}(A_i) \, \| \, \widehat{\bm{L}}]\in\mathbb{R}^{3+100+6},
\end{equation}
where $\|$ denotes the concatenation operator. The complete crystal is then represented by the matrix $\bm{V} = [\bm{v}_1, \dots, \bm{v}_N]^\top \in \mathbb{R}^{N \times (3+100+6)}$, where each row corresponds to an individual atom vector.


\paragraph{Atom Local Environment}
Since there is no unique way to define edges in crystals~\cite{cdvae}, prior studies typically model the local pattern of each atom by considering either its $k$-nearest neighbors~\cite{diffcsp} or all neighbors within a cutoff radius~\cite{mattergen}. In this paper, we construct local atomic environments using the \textit{minimum-distance} rule~\cite{lostop}:
\begin{itemize}
    \item For atom $i$, find its nearest neighbor $j$ and denote their distance by $d_{\min}$.
    \item Retain other neighboring atoms $q$ (up to $K$ neighbors) that satisfy $d_{iq}\leq \xi \cdot d_{\min}$,
\end{itemize}
where $\xi$ is set to 1.1 following standard practice.

\subsection{Diffusion-Based Generation}
The family of diffusion models~\cite{origin, stable} has become one of the most powerful paradigms in generative modeling. In this work, we adopt a standard diffusion framework, DDPM~\cite{ddpm}. In DDPM, the forward process gradually injects Gaussian noise into a sample $\bm{x}$ at each step $s$:
\begin{equation}
\label{forward}
\bm{x}_s = \sqrt{1-\beta_s}\bm{x}_{s-1} + \sqrt{\beta_s}\bm{\epsilon}_s
= \sqrt{\bar{\alpha}_s}\bm{x}_0 + \sqrt{1-\bar{\alpha}_s}\bm{\epsilon}_s,\quad \bm{\epsilon}_s\sim\mathcal{N}(0, \bm{I}),
\end{equation}
where $\{\beta_1,\dots,\beta_S\}$ is a predefined variance schedule, typically satisfying $\beta_1<\dots<\beta_S$, $\alpha_s := 1-\beta_s$, and $\bar{\alpha}_s := \prod_{i=1}^s\alpha_i$. The simplified DDPM objective is to predict the injected noise $\bm{\epsilon}_s$ using a denoising network $\bm{\epsilon}_\theta(\bm{x}_s, s)$:
\begin{equation}
\label{ddpm_objective}
    L_{\mathrm{DDPM}}(\theta) := \mathbb{E}_{s,\bm{x}_0,\bm{\epsilon}}\left[\|\bm{\epsilon}_s-\bm{\epsilon}_\theta(\bm{x}_s, s)\|^2\right].
\end{equation}

\paragraph{Classifier-free Guidance}
In addition to the timestep $s$, one can condition the denoising network on an auxiliary signal $c$, written as $\bm{\epsilon}_\theta(\bm{x}_s, s, c)$, to guide generation. Classifier-free guidance (CFG)~\cite{cfg} enables conditional generation without requiring gradients from an external classifier. By training the same denoising network in both conditional and unconditional modes, CFG estimates the noise at step $s$ as
\begin{equation}
\label{cfg}
   \tilde{\bm{\epsilon}}_s=(1+\omega)\cdot\bm{\epsilon}_\theta(\bm{x}_s, s, c)-\omega\cdot\bm{\epsilon}_\theta(\bm{x}_s, s, \varnothing),
\end{equation}
where $\varnothing$ denotes the unconditional input and $\omega>0$ controls the guidance strength.
\begin{figure*}[t]
  \centering
  \includegraphics[scale=0.2]{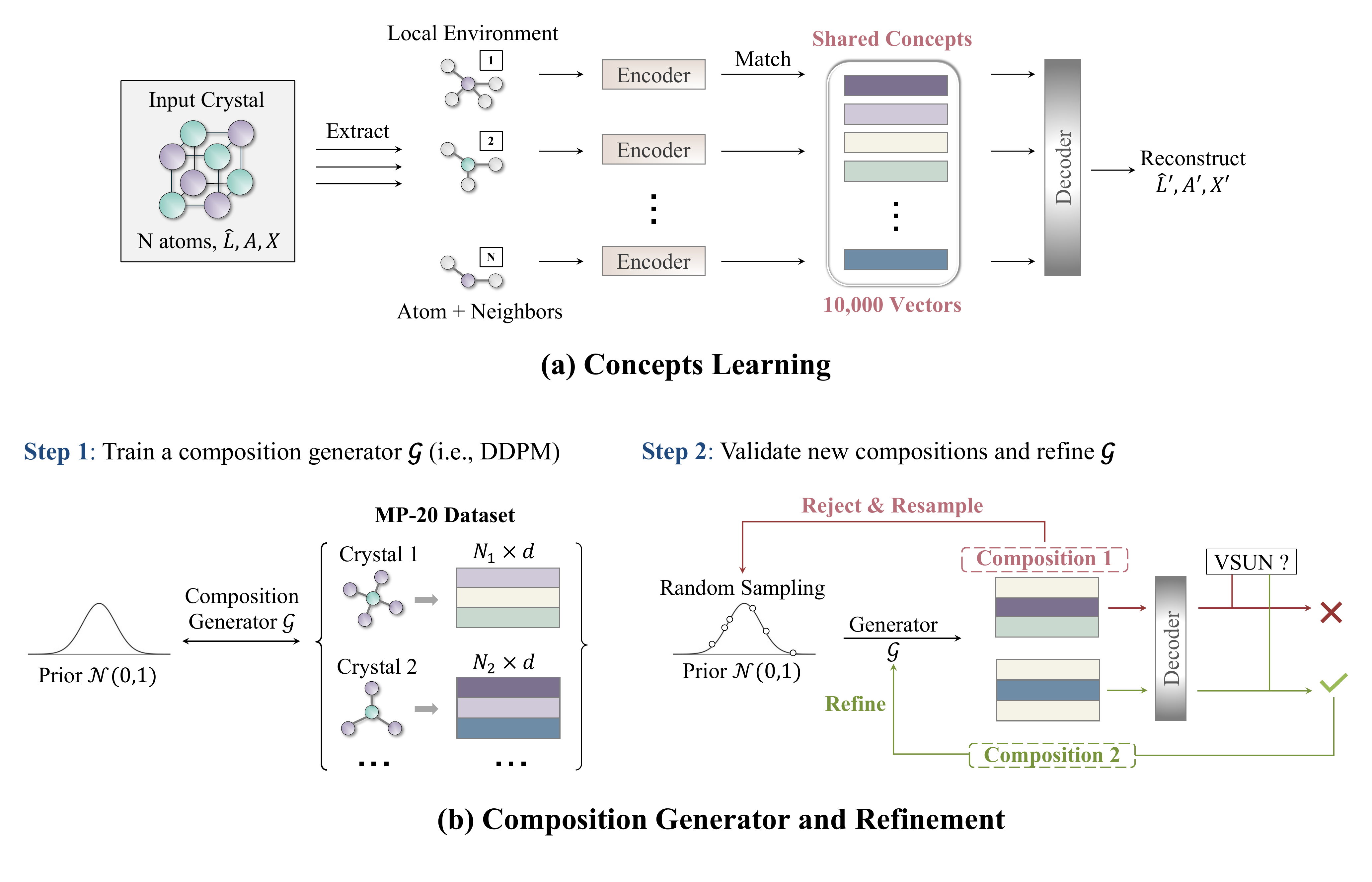}
  \caption{Pipelines for (a) learning a VQ-VAE to extract crystal concepts, and (b) training and refining a composition generator using qualified generated samples.}
  \label{framework}
\end{figure*}
\section{Crystal Concepts and Composition}
\subsection{Concept Extraction}
\label{Concept Extraction}
In this section, we extract crystal concepts from the MP-20 training set~\cite{cdvae}, which contains nearly all experimentally stable materials from the Materials Project~\cite{mp} with at most 20 atoms per unit cell. The training set contains 27,136 crystal structures. We train a VQ-VAE~\cite{vqvae} on this dataset and treat the learned codebook entries as crystal concepts. The overall pipeline is illustrated in Fig.~\ref{framework}(a).

Given a crystal $\mathbf{C}$ with $N$ atoms, we first construct local atomic environments using the minimum-distance rule and collect at most $K$ neighbors for each atom $i\in\mathbf{C}$.\footnote{If atom $i$ has fewer than $K$ neighbors under this rule, we pad the sequence with special tokens until its length reaches $K$.} We then concatenate the atom vectors (Eq.~\eqref{atom_rep}) of atom $i$ and its $K$ neighbors, yielding a matrix $\bm{T}_i\in\mathbb{R}^{(1+K)\times(3+100+6)}$. Next, we feed all such $N$ local atom-neighbor groups $\{\bm{T}_i\}$ from crystal $\mathbf{C}$ into the VQ-VAE:
\begin{equation}
    \label{vqvae}
    \{\bm{z}_i\}=\texttt{Encoder}(\{\bm{T}_i\})\in\mathbb{R}^{N\times d},\quad \{t_i\}=\arg\min_{t}||{\bm{z}_i}-\bm{e}_{t}||^2_2,\quad \bm{A}',\bm{X}',\bm{\widehat{L}}'=\texttt{Decoder}(\{\bm{e}_{t_i}\}),
\end{equation}
where $d$ is the latent dimension, and $\{\bm{e}_t\}_{t=1}^T$ denotes the VQ-VAE codebook containing $T$ codes, which we interpret as crystal concepts. In our experiments, we set $T=10{,}000$.

Both \texttt{Encoder} and \texttt{Decoder} are implemented using vanilla Transformers~\cite{vaswani2017attention}, but with different interaction scopes. The \texttt{Encoder} operates on each $\bm{T}_i$ independently, such that each atom interacts only with its own local neighborhood. This design prevents the latent representations $\{\bm{z}_i\}$ from entangling distinct local patterns. In contrast, the \texttt{Decoder} processes all $N$ atoms jointly to reconstruct the full crystal, i.e., $\bm{A}'$, $\bm{X}'$, and $\widehat{\bm{L}}'$. The quantization step in Eq.~\eqref{vqvae} matches each $\bm{z}_i$ to its nearest codebook entry $\bm{e}_{t_i}$, and replaces $\{\bm{z}_i\}$ with the corresponding concept vectors for reconstruction. The training objective consists of
\begin{equation}
\begin{aligned}
\label{vqvae_loss}
    &\mathcal{L}_{\bm{A}}=\frac{1}{N}\sum_i\mathrm{CrossEntropy}(\bm{A}_i,\bm{A}_i'),\quad \mathcal{L}_{\bm{X}}=\frac{1}{N}\sum_i||\bm{X}_i-\bm{X}_i'||^2_2,\\ &\mathcal{L}_{\bm{L}}=||\bm{\widehat{L}}-\bm{\widehat{L}}'||^2_2,\quad \mathcal{L}_{reg}=\frac{1}{N}\sum_i||\texttt{sg}[\bm{z}_i]-\bm{e}_{t_i}||^2+||\bm{z}_i-\texttt{sg}[\bm{e}_{t_i}]||^2,
\end{aligned}
\end{equation}
where $\texttt{sg}[\cdot]$ is stop-gradient operation. More training details are given in Appendix~\ref{vqvae_training}.

\subsection{Concept Composition}
In this section, we describe how to generate new concept compositions, validate the generated outputs, and use the qualified samples to refine the composition generator. After refinement, the generator is used to sample multiple concept compositions, which then guide the downstream base generative model. These two stages are illustrated in Fig.~\ref{framework}(b).

\paragraph{Training a Composition Generator.}
Prior work~\cite{conceptmix, yu2023skill} typically composes concepts through random sampling, which is inefficient for crystal generation due to stringent physical constraints. Instead, we introduce a composition generator $\mathcal{G}$ to perform concept composition directly.

Specifically, for each crystal $\mathbf{C}$ with $N_C$ atoms in MP-20, we construct its $N_C$ local environments and feed them into the encoder to obtain $\bm{Z}_C\in\mathbb{R}^{N_C\times d}$. The generator $\mathcal{G}$, implemented as a DDPM, is trained to learn a mapping from Gaussian noise to the set of pre-quantization latent matrices $\{\bm{Z}_C\}$ of all crystals in MP-20. After training, $\mathcal{G}$ generates new latent matrices $\{\bm{Z}^g\}$, which are then quantized using Eq.~\eqref{vqvae} to obtain $\{\bm{E}^g\}$. Each matrix $\bm{E}^{g} = [\bm{e}_{t_1}, \dots, \bm{e}_{t_{N_g}}]^\top \in \mathbb{R}^{N_{g} \times d}$
represents a composition of crystal concepts, where $N_{g}$ is the number of atoms in the generated crystal.

\paragraph{Validating and Refining the Generator.}
Whether concepts are selected randomly or generated by a composition model, only valid concept compositions can support meaningful downstream crystal generation. For example, \cite{conceptmix} presented an invalid composition such as ``a triangle-shaped person.'' We therefore need criteria for filtering high-quality compositions from the generated candidates.

In materials discovery, four criteria are particularly important in practice: validity (V), stability (S), uniqueness (U), and novelty (N). Among them, validity and uniqueness can be checked directly from the generated crystals themselves, whereas stability and novelty require comparison against an external reference set containing known crystals. However, during model training, we avoid relying on such external references for two reasons: (1) there is no single unified reference set across all experimental settings, and (2) using external data during training would lead to an unfair comparison with other baselines. Based on these considerations, we use all crystals available during model training as the reference set, and evaluate stability and novelty relative to it. Our underlying assumption is that these training crystals are stable regardless of the final evaluation reference. Moreover, a generated crystal with comparable formation energy under the same chemical system is likely to be stable as well.

After training the composition generator $\mathcal{G}$, we use it to produce a pool of new compositions $\{\bm{E}^g\}$, and retain only those whose decoded crystals satisfy the V.S.U.N. criteria with respect to the training set. We then fine-tune $\mathcal{G}$ on these validated compositions as a refinement step, following the same training protocol as the previous stage.

\paragraph{Composition-based Generative Model.}
Once refined, the generator $\mathcal{G}$ produces new compositions $\{\bm{E}^g\}$ that enable the downstream generative model to explore novel crystal structures in a controllable manner. In this work, we implement the composition-based model using a Denoising Diffusion Probabilistic Model (DDPM) with a vanilla Transformer architecture conditioned on these compositions. To validate the utility of the extracted concepts, we evaluate whether conditioning on the final $\{\bm{E}^g\}$ enhances generation quality relative to both unconditional models and baselines.


\subsection{Transformers as Backbones}
\begin{figure*}[ht]
  \centering
  \includegraphics[scale=0.25]{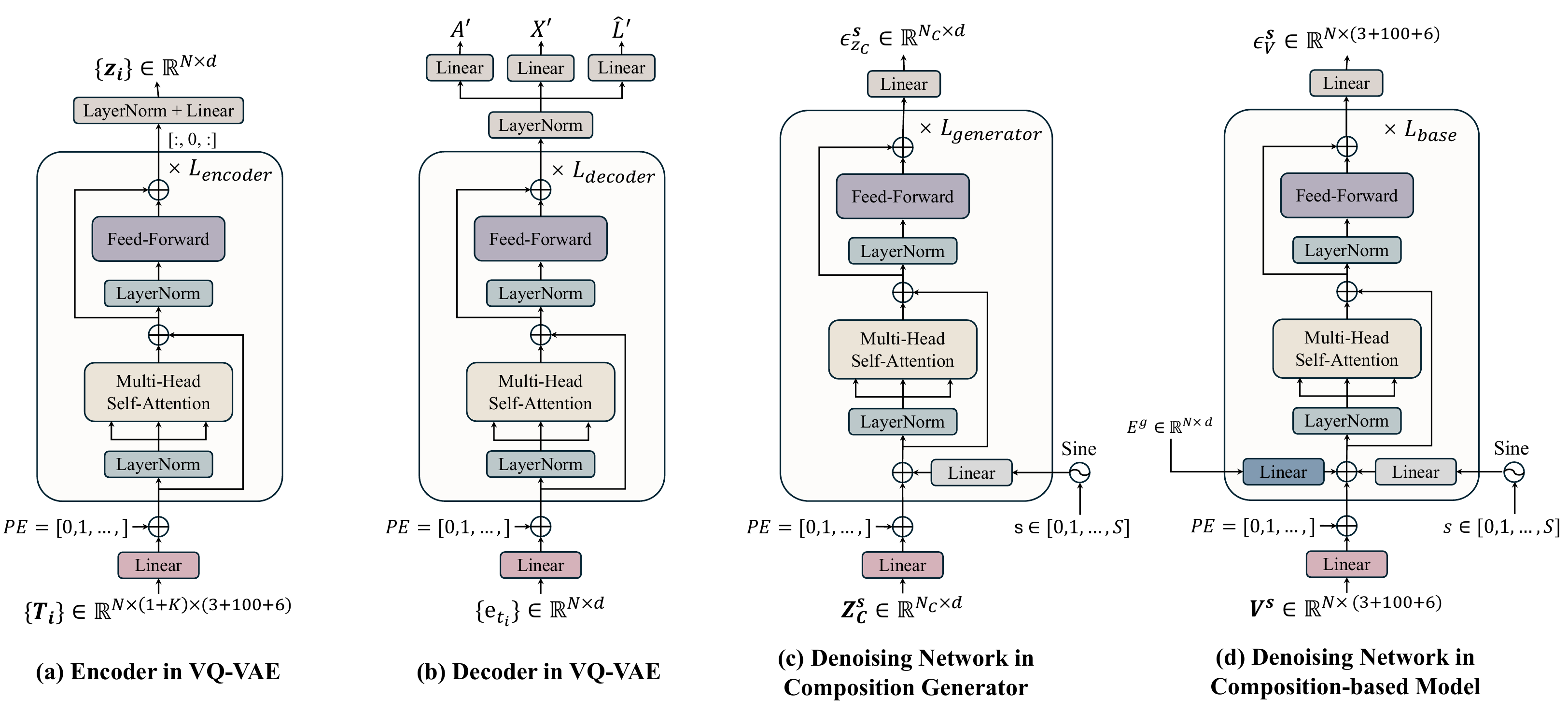}
  \caption{Transformer architectures used in (a) the encoder and (b) the decoder of the VQ-VAE, as well as the denoising networks in (c) the composition generator and (d) the composition-based model.}
  \label{transformer}
\end{figure*}
Throughout concept extraction and composition, we use backbone models as the encoder and decoder in the VQ-VAE, and as the denoising networks in both the composition generator and the base generative model. For all four cases, we adopt the vanilla Transformer architecture~\cite{vaswani2017attention} as our backbone, incorporating several key modifications. The detailed designs are shown in Fig.~\ref{transformer}, and more detailed introduction can be found in Appendix~\ref{related work}. The first modification is the use of pre-normalization before multi-head attention and feed-forward layers~\cite{pre-norm}. We also introduce a simple positional encoding scheme and a condition incorporation strategy.

\paragraph{Positional Encoding (PE).}
There is no widely accepted canonical ordering of atoms in a crystal. However, as sequence-based models, Transformers require an ordering scheme to function effectively. In this paper, we order atoms according to two principles: (1) atoms with smaller atomic numbers are placed before those with larger atomic numbers, and (2) among atoms of the same type, we compare their coordinates sequentially along the x-, y-, and z-axes, placing atoms with smaller coordinates first. After this ordering procedure, each atom is assigned a learnable position vector according to its order. These position vectors are shared across all crystals.

\paragraph{Condition Incorporation.}
In the denoising networks, additional conditions are incorporated as part of the input, including the diffusion timestep $s$ and the generated concept composition $\bm{E}^g$. For timestep $s$, we apply a linear layer after sinusoidal encoding to obtain a time embedding. For composition $\bm{E}^g$, we use another linear layer to obtain a composition embedding. These condition embeddings are then added to the input at the beginning of each Transformer layer.
\section{Concept Interpretation}
\label{interpret}
After training the VQ-VAE, we analyze the optimized 10,000 discrete codes as crystal concepts, since each code represents a cluster of similar local atomic environments. In this section, we interpret these learned concepts from both local and global perspectives.

\begin{figure*}[ht]
  \centering
  \includegraphics[scale=0.18]{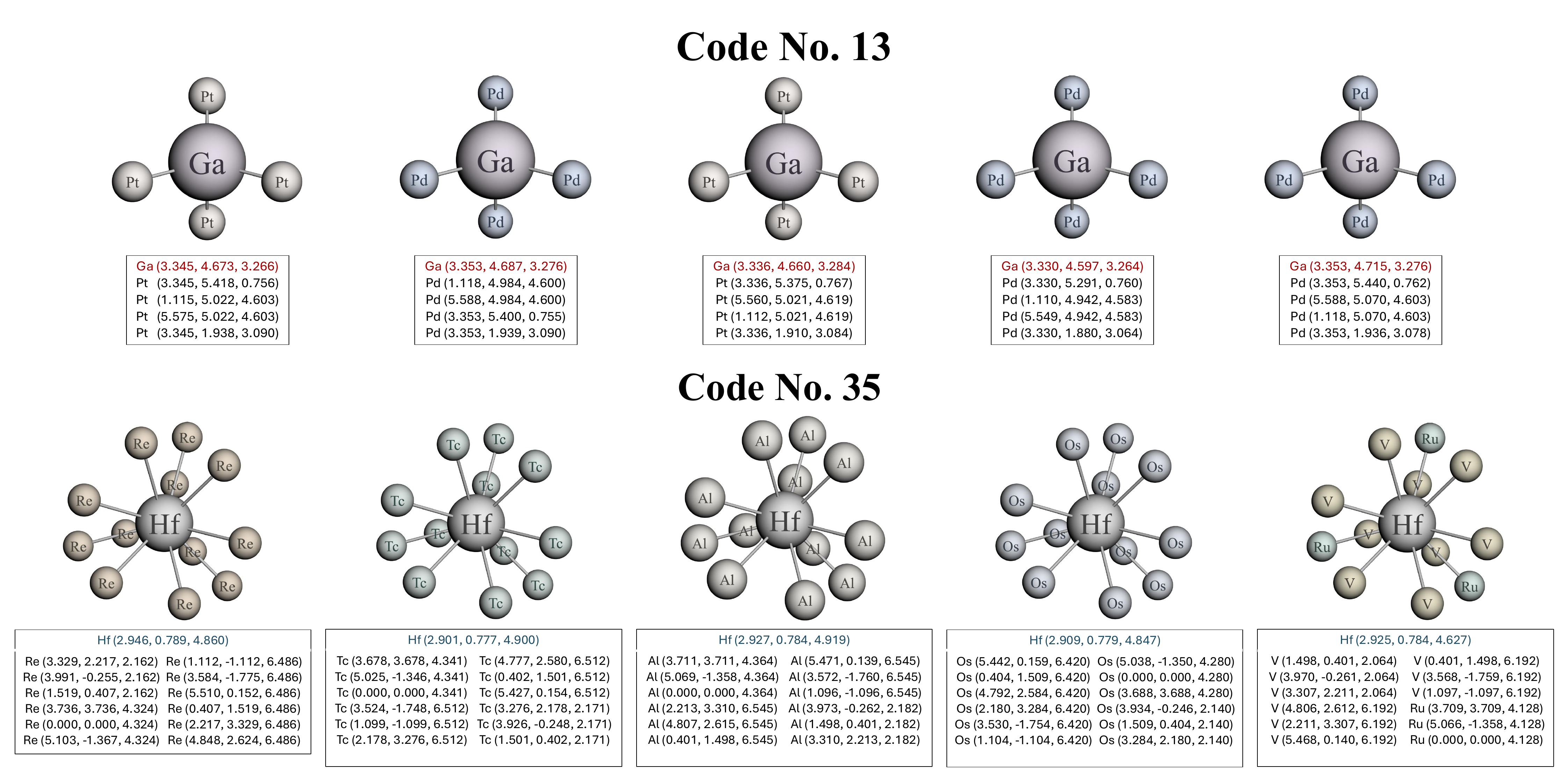}
  \caption{Illustrations of two learned concepts represented by VQ-VAE codes.}
  \label{local}
\end{figure*}

\paragraph{Local view.}
We first examine what local information is encoded by the learned concepts. Specifically, for a given code $\bm{e}_t$, we collect all local environments $\{\bm{T}_i\}\in\mathbb{R}^{(1+K)\times(3+100+6)}$ whose latent representations $\{\bm{z}_i\}$ are assigned to $\bm{e}_t$ during the quantization step in Eq.~\eqref{vqvae}. We then compute the distances $\{d_i^{\bm{e}_t}=\|\bm{z}_i-\bm{e}_t\|_2^2\}$ and retrieve the top-5 local environments with the smallest distances, i.e., those that best match concept $\bm{e}_t$. Finally, we visualize their atomic coordinates and species. Two examples are shown in Fig.~\ref{local}, from which we draw the following observations:
\begin{itemize}
    \item The most representative local environments of the same concept can originate from different chemical systems. However, they exhibit similar 3D structures, suggesting that the learned concepts primarily capture geometric patterns.
    \item Elements with similar physical properties tend to share the same concepts. In Fig.~\ref{local}, the element pairs Pt--Pd, Re--Tc, and Ru--Os each belong to the same group in the periodic table, and they co-occur within the same concepts as neighbors of the target atom.
\end{itemize}
It should be noted that not all of the 10,000 learned concepts exhibit such visually distinct patterns. This is anticipated, as the concepts are learned in an unsupervised manner without the imposition of human-defined chemical or structural priors. Consequently, the interpretable patterns that emerge from the learned codebook serve as compelling evidence that the model captures non-trivial regularities within local crystal environments. We provide additional local-view visualizations in Appendix~\ref{more interpret}.

\paragraph{Global view.}
Space groups are commonly used to describe the 230 possible types of global symmetry in crystals. These groups can be further categorized into six crystal families. Table~\ref{family} summarizes the relationship between space groups and crystal families. In this part, we investigate how the learned local concepts relate to these global structural properties.
\begin{table}[ht]
\caption{Lattice shape and included space groups of crystal families~\cite{diffcsppp}, where $a,b,c$ and $\alpha,\beta,\gamma$ denote the lengths and angles of the lattice base, respectively.}
\label{family}
  \vskip 0.1in
  \centering
  \renewcommand\arraystretch{1.2}
    \resizebox{\textwidth}{!}
{
    \begin{tabular}{cccccc|c}
        \toprule[1.2pt]
        Crystal Family & Triclinic & Monoclinic & Orthorhombic & Tetragonal & Cubic & Hexagonal\\
        \midrule
        Space Group No. & 1$\sim$2 & 3$\sim$15 & 16$\sim$74 & 75$\sim$142 & 195$\sim$230 & 143$\sim$194\\
        Lattice Shape & No Constraint & $\alpha=\gamma=90^\circ$ & $\alpha=\beta=\gamma=90^\circ$ & \makecell[c]{$\alpha=\beta=\gamma=90^\circ$\\$a=b$} & \makecell[c]{$\alpha=\beta=\gamma=90^\circ$\\$a=b=c$} & \makecell[c]{$\alpha=\beta=90^\circ$,\\$\gamma=120^\circ$, $a=b$}\\
        \bottomrule[1.2pt]
    \end{tabular}}
\end{table}

\begin{wrapfigure}[13]{r}{0.5\textwidth}
\vspace{-1.5em}
    \centering
      \includegraphics[scale=0.4]{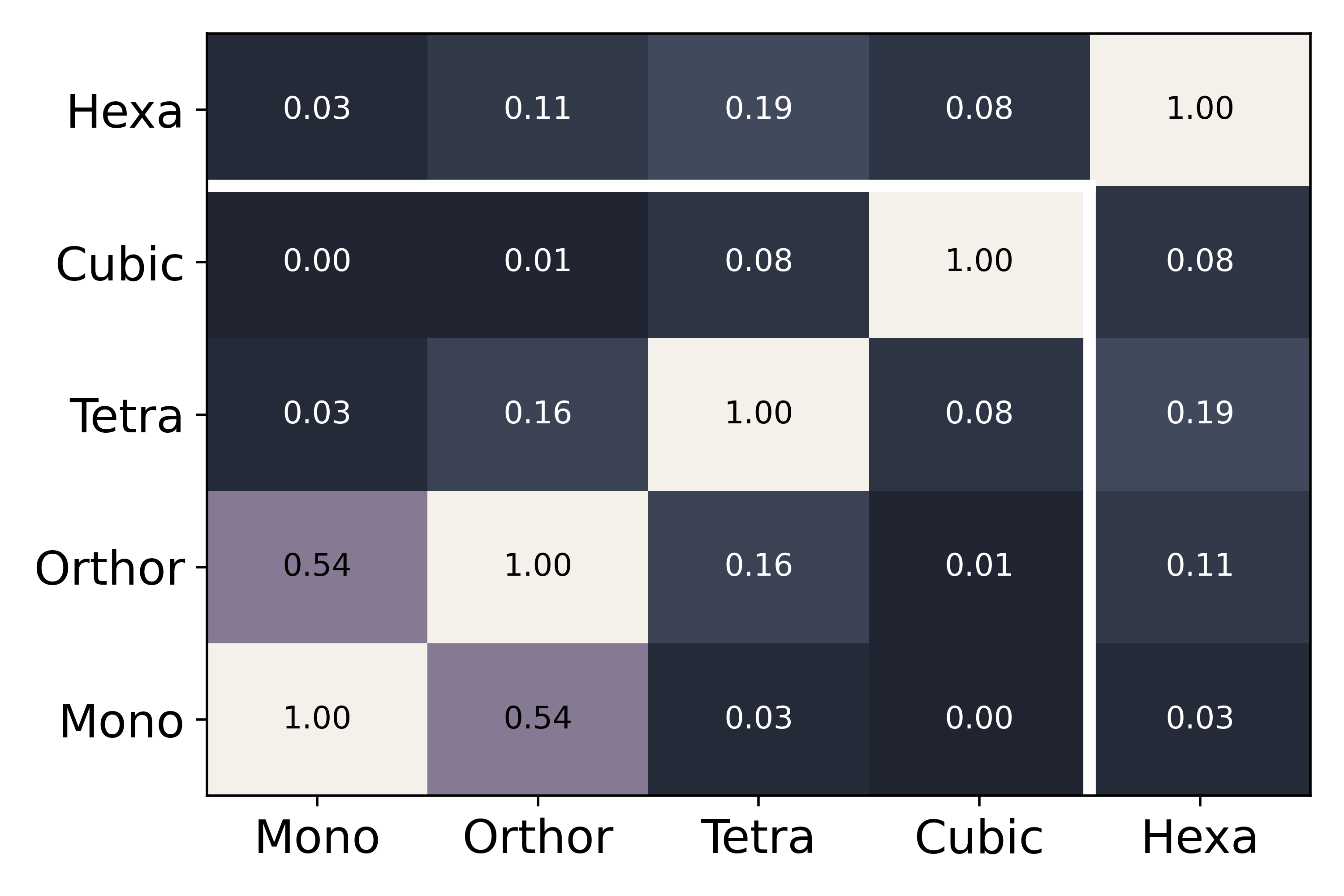}
      \caption{Cosine similarities of concept selection across five crystal families.}
      \label{global_view}
\end{wrapfigure}
\textbf{a. Crystal Families Select Concepts.}
We first categorize all MP-20 crystals according to their crystal families. For each family $\mathcal{F}$, we construct a probability vector $\bm{p}^{\mathcal{F}}=(p_{e_1}^{\mathcal{F}},\cdots,p_{e_T}^{\mathcal{F}})\in\mathbb{R}^T$, where $p_{e_j}^{\mathcal{F}}$ denotes the frequency with which the $j$-th code is selected among all crystals in family $\mathcal{F}$. We then compute pairwise cosine similarities among the vectors $\{\bm{p}^{\mathcal{F}}\}$, as shown in Fig.~\ref{global_view}. In the figure, we exclude the comparison against \texttt{Triclinic} family, since this trivial family does not have obvious global pattern in Table~\ref{family}. From the figure, we make two observations:

\quad (1) Table~\ref{family} shows that \texttt{Cubic} $\rightarrow$ \texttt{Tetragonal} $\rightarrow$ \texttt{Orthorhombic} $\rightarrow$ \texttt{Monoclinic} forms a symmetry-reduction hierarchy, where moving downward corresponds to progressively relaxed lattice constraints. This hierarchy is well aligned with Fig.~\ref{global_view}, in which adjacent families exhibit higher similarity in concept selection. For example, \texttt{Orthorhombic} has higher similarity to \texttt{Monoclinic} and \texttt{Tetragonal}, and these three families are adjacent in the hierarchy.

\quad (2) Although \texttt{Hexagonal} is not part of this chain, its lattice differs from \texttt{Tetragonal} mainly in the $\gamma$ angle. Consequently, \texttt{Hexagonal} crystals exhibit concept preference distributions that are more similar to those of \texttt{Tetragonal} crystals than to those of other families in Fig.~\ref{global_view}.

\begin{wraptable}[7]{r}{0.5\textwidth}
\vspace{-1.em}
\caption{Prediction on space groups and crystal families based on concepts.}
  \label{prediction}
  \small
  \centering
  \begin{tabular}{ccc}
    \toprule[1.2pt]
    \% &  Space Group   &  Crystal Family \\
    \midrule 
    Accuracy  & 67.11 & 77.37\\
    \bottomrule[1.2pt]
  \end{tabular}
\end{wraptable}
\textbf{b. Concepts Predict Groups And Families.}
Given only the selected concepts of a crystal, we predict its space group and crystal family to further strengthen their connection. Specifically, for a crystal $\mathcal{C}$ with $N$ selected concepts $\{\bm{e}_{t_1}, \ldots, \bm{e}_{t_N}\}$, we concatenate these concept embeddings into a matrix $\bm{E}=[\bm{e}_{t_1},\cdots,\bm{e}_{t_N}]^\top$.
We then feed $E$ into an eight-layer Transformer with hidden dimension 512. The output representations are aggregated by mean pooling and passed through a two-layer multi-layer perceptron (MLP) to separately predict 230 space groups and 6 crystal families. As shown in Table~\ref{prediction}, the selected concepts alone achieve 67.11\% accuracy for space-group prediction and 77.37\% accuracy for crystal-family prediction. These results indicate that although the concepts are learned from local atomic environments, their compositions retain substantial information about global crystal symmetry.

\section{Concept Generalization}
In this section, we evaluate the extent to which concept composition enhances generative performance, with a specific focus on structural novelty. Furthermore, we assess the generalizability of concepts learned from MP-20 to Alex-MP-20~\cite{mattergen}, a dataset characterized by a distinct distribution of crystal structures. Implementation details are provided in Appendix~\ref{Implementation Details}.

\subsection{Benchmark on MP-20}
\begin{table}[h]
  \caption{Evaluate 10,000 generative crystals from the models trained on MP-20. Reference is MP-2023. ($\dag$ means the authors did not provide model checkpoints. We report their results according to our reproduction.)}
  \vskip 0.1in
  \label{mp20_re}
  \small
  \centering
  \renewcommand\arraystretch{1.2}
  \resizebox{1\columnwidth}{!}
{
  \begin{tabular}{ccccc|cccc}
    \toprule[1.2pt]
           \%       &  Validity (V)   &  Stability (S)  &  Uniqueness (U) &  Novelty (N) & \makecell{S.U.N\\(MatterSim)} & \makecell{V.S.U.N\\(MatterSim)} & \makecell{S.U.N\\(DFT)} & \makecell{V.S.U.N\\(DFT)} \\
    \midrule 
    CDVAE$^\dag$~\cite{cdvae} &85.3$\pm$1.3&29.9$\pm$1.2&99.8$\pm$0.2&\textbf{96.5$\pm$0.6}&27.0$\pm$1.4&22.8$\pm$1.1&-&-\\
    DiffCSP~\cite{diffcsp} &83.1$\pm$1.3&45.9$\pm$1.8&99.5$\pm$0.1&83.6$\pm$0.6&30.9$\pm$1.9&25.6$\pm$1.3&-&-\\
    DiffCSP++~\cite{diffcsppp} &85.0$\pm$0.9&39.7$\pm$2.0&99.8$\pm$0.1&82.8$\pm$1.0&23.9$\pm$1.3&20.3$\pm$1.2&-&-\\
    FlowMM$^\dag$~\cite{flowmm} &81.8$\pm$0.8&40.8$\pm$2.0&99.8$\pm$0.0&83.1$\pm$1.1&25.3$\pm$1.7&21.0$\pm$1.3&-&-\\
    FlowLLM$^\dag$~\cite{Flowllm} &83.8$\pm$0.7&36.5$\pm$1.5&98.2$\pm$0.5&86.4$\pm$1.5&25.1$\pm$0.6&21.3$\pm$0.8&-&-\\
    SymmCD$^\dag$~\cite{symmcd} &75.5$\pm$1.5&34.7$\pm$1.4&99.7$\pm$0.1&85.1$\pm$1.5&19.0$\pm$1.0&15.7$\pm$0.8&-&-\\
    ADiT~\cite{adit} &89.7$\pm$1.0&69.5$\pm$1.1&99.1$\pm$0.2&58.9$\pm$0.9&30.3$\pm$0.9&27.2$\pm$1.3&-&-\\
    CrysLLMGen (7B)$^\dag$~\cite{crysllmgen} &88.1$\pm$0.9&35.1$\pm$1.9&98.4$\pm$0.3&87.4$\pm$1.1&22.9$\pm$0.9&20.4$\pm$0.9&-&-\\
    SGEquiDiff~\cite{sge} &85.1$\pm$1.0&46.5$\pm$1.9&98.9$\pm$0.4&74.8$\pm$1.2&23.6$\pm$0.9&20.0$\pm$1.0&-&-\\
MatterGen-MP~\cite{mattergen} &84.2$\pm$1.2&47.0$\pm$1.1&\textbf{99.8$\pm$0.1}&86.6$\pm$1.0&34.7$\pm$1.3&29.2$\pm$0.9&30.8&26.4\\
    \midrule
    Base model &82.1$\pm$1.5&\textbf{48.4$\pm$1.4}&99.6$\pm$0.0&78.0$\pm$1.6&27.8$\pm$1.1&23.3$\pm$1.3&-&-\\
    w/ composition &\textbf{93.9$\pm$0.4}&46.9$\pm$1.0&98.8$\pm$0.3&90.0$\pm$0.5&\textbf{36.4$\pm$1.5}&\textbf{35.7$\pm$1.7}&\textbf{31.8}&\textbf{31.1}\\
    \bottomrule[1.2pt]
  \end{tabular}
  }
\end{table}
\paragraph{Evaluation Setting.}
We repeat sampling 10 times and collect 1,000 crystals in each run. We then report the mean and standard deviation of \texttt{validity (V)}, \texttt{stability (S)}, \texttt{uniqueness (U)}, and \texttt{novelty (N)}. We also report \texttt{S.U.N} and \texttt{V.S.U.N}.\footnote{Only valid crystals are meaningful for evaluating the remaining properties.} During evaluation, structure relaxation and energy prediction are performed using MatterSim-v1-1M~\cite{mattersim}.
Following prior work~\cite{flowmm}, we use MP-2023 as the reference dataset and regard crystals with energy above hull below 0.1 eV/atom as stable. We further perform density functional theory (DFT) calculations~\cite{perdew1996pbegga} to provide a more rigorous stability assessment, with details provided in Appendix~\ref{dft}. Due to the high computational cost, we only perform DFT on 1,000 crystals sampled from both our model and the strongest baseline.

\paragraph{Baselines.}
We train our composition-based generative model (DDPM + Transformer) on MP-20, and further condition it on the compositions generated by $\mathcal{G}$. We compare these two variants against state-of-the-art baselines, including CDVAE~\cite{cdvae}, DiffCSP~\cite{diffcsp}, DiffCSP++~\cite{diffcsppp}, FlowMM~\cite{flowmm}, FlowLLM~\cite{Flowllm}, SymmCD~\cite{symmcd}, ADiT~\cite{adit}, CrysLLMGen (7B)~\cite{crysllmgen}, SGEquiDiff~\cite{sge} and MatterGen~\cite{mattergen}.

\paragraph{Results.}
As shown in Table~\ref{mp20_re}, incorporating concept compositions makes our composition-base model the best-performing method, achieving improvements of 17.8\% in \texttt{V.S.U.N} using DFT over MatterGen, the strongest baseline. In particular, concept composition preserves stability while substantially improving novelty. These results show that compositional generation can effectively guide structured exploration beyond the training distribution.

\subsection{Generalization to Alex-MP-20}
\begin{table}[h]
  \caption{Evaluate VQ-VAE reconstruction on MP-20 and Alex-Mp-20}
  \vskip 0.1in
  \label{vqvae_recon}
  \footnotesize
  \centering
  \begin{tabular*}{0.85\textwidth}{@{\extracolsep{\fill}}ccccc}
    \toprule[1.2pt]
           \%       &  MP-20 training &  MP-20 validation &  MP-20 test &  Alex-MP-20\\
    \midrule 
    Reconstruction ratio & 96.2 & 82.7 & 83.0 & 61.1\\
    \bottomrule[1.2pt]
  \end{tabular*}
  
\end{table}
\paragraph{VQ-VAE reconstruction.}
After training the VQ-VAE, we freeze the learned codebook and pass the crystals through the model, then evaluate whether each reconstructed crystal matches the input.\footnote{We use the \texttt{StructureMatcher} class in \texttt{pymatgen}~\cite{pymatgen} with thresholds \texttt{stol}=0.5, \texttt{angle\_tol}=10, and \texttt{ltol}=0.3.} As shown in Table~\ref{vqvae_recon}, these concepts can still successfully reconstruct 61\% of crystals from Alex-MP-20, indicating that the learned concepts can represent unseen crystals from another distribution to a meaningful extent.

\begin{table}[h]
  \caption{Evaluate 10,000 generative crystals from the models trained on Alex-MP-20. Reference is MP2020correction provided by MatterGen~\cite{mattergen}.}
  \label{alexmp20_re}
  \vskip 0.1in
  \small
  \centering
  \renewcommand\arraystretch{1.2}
  \resizebox{1\columnwidth}{!}
{
  \begin{tabular}{ccccc|cccc}
    \toprule[1.2pt]
    \% &  Validity (V)   &  Stability (S)  &  Uniqueness (U) &  Novelty (N) & \makecell{S.U.N\\(MatterSim)} & \makecell{V.S.U.N\\(MatterSim)} & \makecell{S.U.N\\(DFT)} & \makecell{V.S.U.N\\(DFT)}\\
    \midrule 
    MatterGen &85.7$\pm$1.2&\textbf{73.1$\pm$1.0}&\textbf{99.9$\pm$0.0}&67.9$\pm$1.4&43.6$\pm$1.2&37.4$\pm$1.4&40.1&35.1\\
    \midrule
    Base model &87.4$\pm$0.9&69.0$\pm$0.9&\textbf{99.9$\pm$0.0}&61.8$\pm$1.2&33.3$\pm$1.4&28.8$\pm$1.0&-&-\\
    w/ composition &\textbf{94.8$\pm$0.8}&59.1$\pm$1.6&99.4$\pm$0.1&\textbf{85.9$\pm$0.9}&\textbf{46.9$\pm$1.4}&\textbf{43.7$\pm$1.5}&\textbf{40.6}&\textbf{40.2}\\
    \bottomrule[1.2pt]
  \end{tabular}
  }
\end{table}
\paragraph{Generation on Alex-MP-20.}
We repeat the same experimental setting for MatterGen and for our composition-based model trained on Alex-MP-20. Notably, the compositions are still formed by recombining concepts learned from MP-20, while their validation is performed using the Alex-MP-20 training set as the reference. As shown in Table~\ref{alexmp20_re}, incorporating concept compositions leads to a decrease in stability. This reduction may suggest that concepts learned from MP-20 do not fully capture the newly emerging patterns in Alex-MP-20. Nevertheless, our primary metrics of interest, \texttt{V.S.U.N.}, still improve by 14.5\% using DFT, due to the substantial gain in novelty brought by compositional generation.

\section{Concept Verification}
\begin{figure*}[ht]
  \centering
  \includegraphics[scale=0.18]{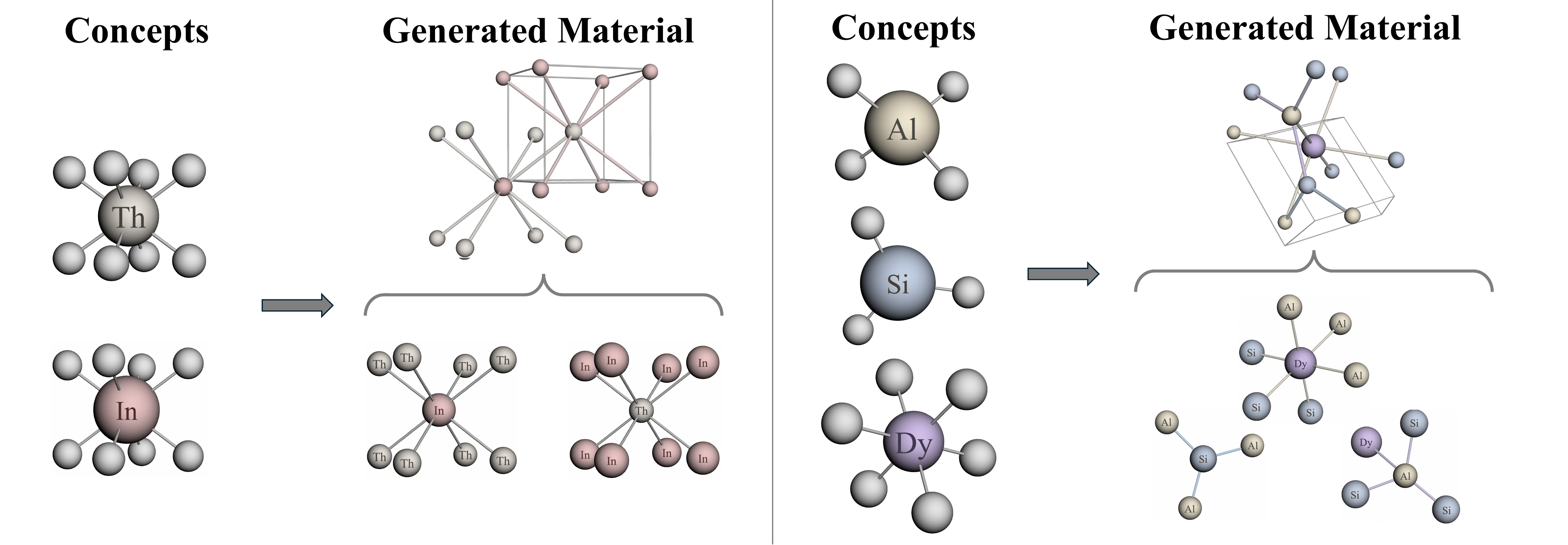}
  \caption{Illustrations of two generated crystals together with their supporting concepts.}
  \label{verify}
\end{figure*}
Finally, we verify whether the generated crystals indeed contain the composed concepts, i.e., whether the generation process reflects genuine compositional generation. Fig.~\ref{verify} shows two generated crystals together with their supporting concepts. The results demonstrate that the base model can effectively understand the extracted concepts and incorporate them into the final generated structures.
\section{Conclusion}
\label{conclusion}
We introduced a compositional framework that uses VQ-VAE-learned concepts to drive \textit{de novo} crystal generation. By recombining these reusable primitives, our model enables a more systematic exploration of structural novelty. Experiments on MP-20 and Alex-MP-20 validate the effectiveness of this approach in producing high-quality, novel candidates.

\textbf{Limitations and broader impact.}
While not all learned concepts are currently human-interpretable, our framework addresses a critical gap in controlling the generative process. This work provides a new paradigm for moving beyond black-box models toward more interpretable and controllable materials discovery.
\begin{ack}
XB is supported by MOE AcRF T1 Grant ID 251RES2423 and NRF AI4SCT Grant ID 20250024. KSN acknowledges support by the National Research Foundation, Singapore under its AI Singapore Programme (AISG Award No: AISG3-RP-2022-028), by the Ministry of Education, Singapore under Research Centre of Excellence award to the Institute for Functional Intelligent Materials, I-FIM (project No. EDUNC-33-18-279-V12), 
by the MAT-GDT Program at A*STAR via the AME Programmatic Fund by the Agency for Science, Technology and Research under Grant No. M24N4b0034 
and by the Tier 3 program (MOE-MOET32024-0001). The computational work for this article was partially performed on resources of the National Supercomputing Centre, Singapore (https://www.nscc.sg).
\end{ack}

\newpage
\bibliographystyle{plain}
\bibliography{bib}

\newpage
\appendix
\section{Visualizing More Local Patterns encoded in Concepts}
\label{more interpret}
In Section~\ref{interpret}, we visualize the top-5 local atomic environments associated with two learned concepts. Here, we provide additional visualizations for concepts with different numbers of neighboring atoms.

\begin{figure}[ht]
    \begin{minipage}{0.75\textwidth}
        \includegraphics[scale=0.18]{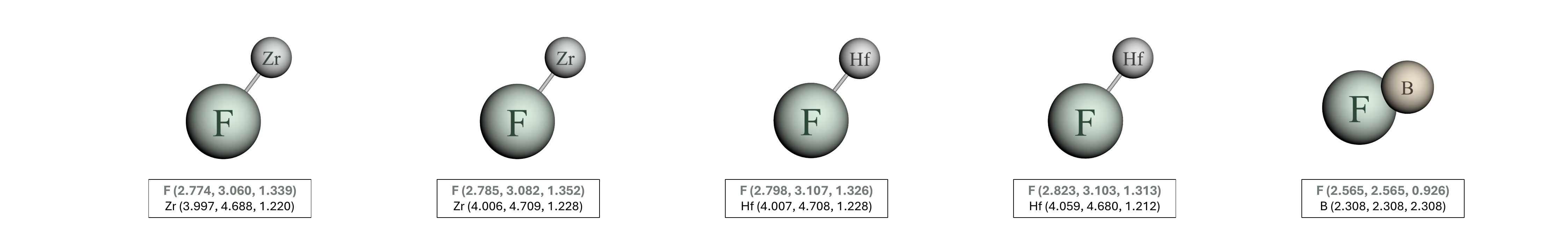}
        \text{\texttt{Zr} and \texttt{Hf} are group-IV transition metals.}
    \end{minipage}

    \vspace{1.5em}

    \begin{minipage}{0.75\textwidth}
  \includegraphics[scale=0.18]{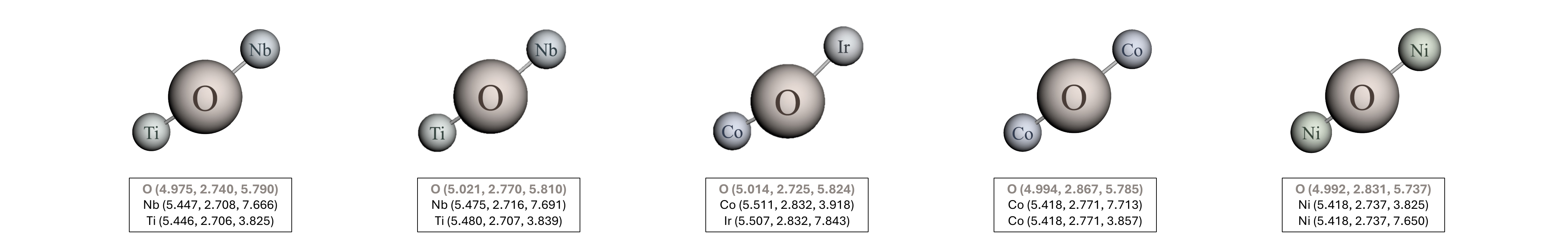}
        \text{\texttt{Co, Ir, Ni, Nb}, and \texttt{Ti} are transition metals, with \texttt{Co--Ir--Ni} and \texttt{Ti--Nb} reflecting chemically related groups.}
        
    \end{minipage}

    \vspace{1.5em}

    \begin{minipage}{0.75\textwidth}
  \includegraphics[scale=0.18]{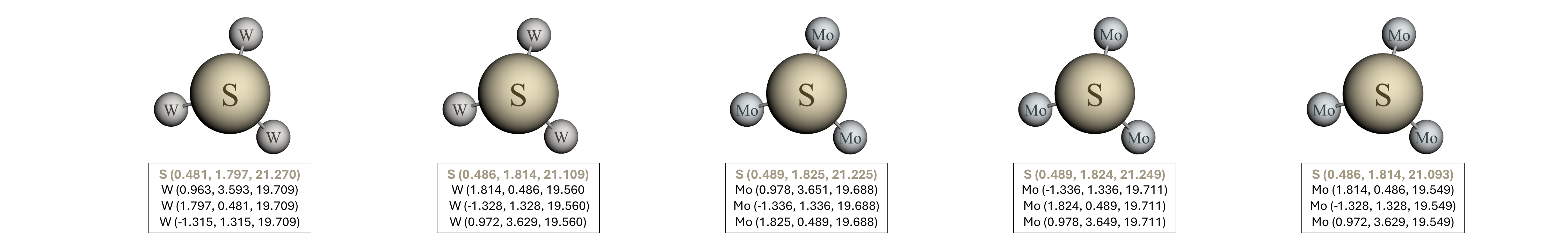}
        \text{\texttt{W} and \texttt{Mo} are group-VI transition metals.}
        
    \end{minipage}

    \vspace{1.5em}

    \begin{minipage}{0.75\textwidth}
  \includegraphics[scale=0.18]{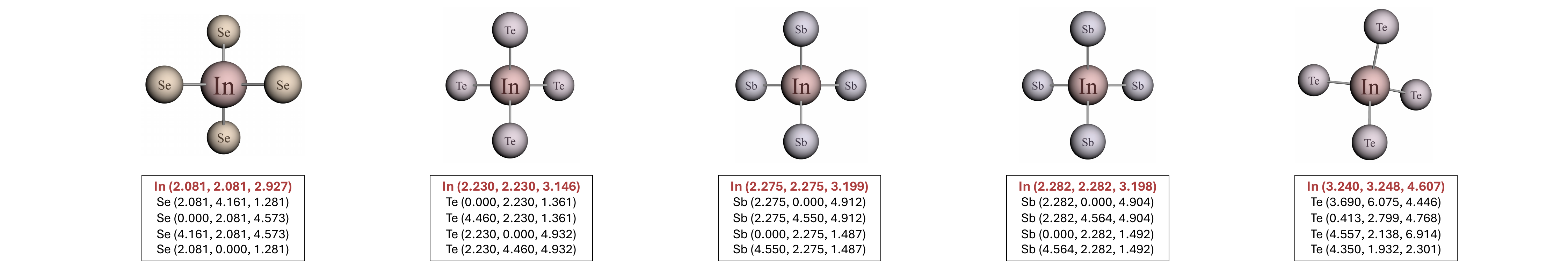}
        \text{\texttt{Se} and \texttt{Te} are group-VI chalcogens, while \texttt{Sb} is adjacent in the periodic table.}
        
    \end{minipage}
    
    \vspace{1.5em}

    \begin{minipage}{0.75\textwidth}
  \includegraphics[scale=0.18]{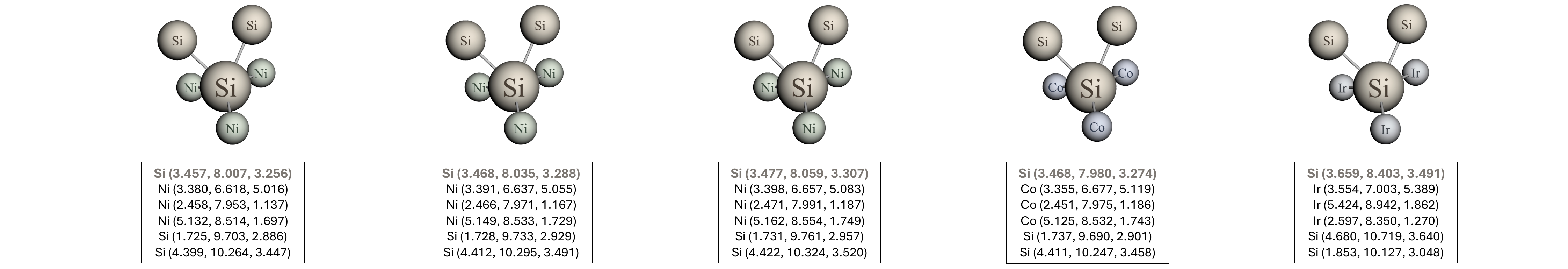}
        \text{Element \texttt{Co} and \texttt{Ir} belong to the same group. \texttt{Co, Ir}, and \texttt{Ni} are transition metals from neighboring groups.}
    \end{minipage}
    \vspace{1.5em}

    \begin{minipage}{0.75\textwidth}
  \includegraphics[scale=0.18]{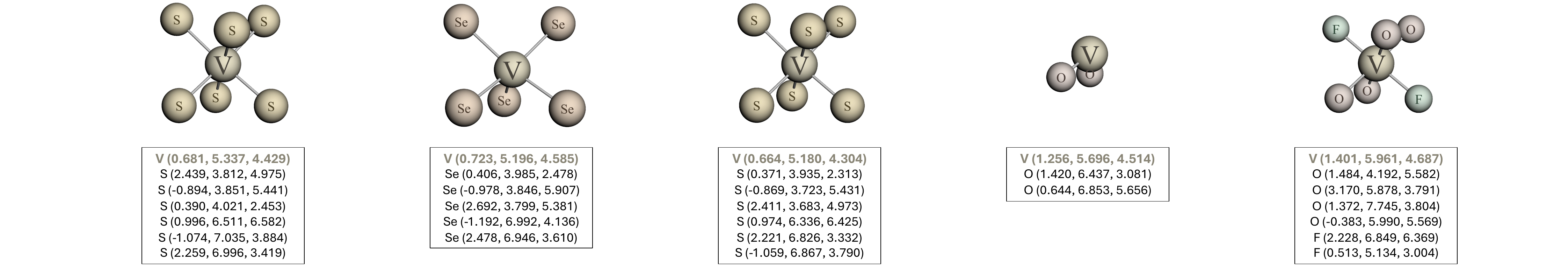}
        \text{\texttt{O, S}, and \texttt{Se} are group-VI chalcogens, while \texttt{F} belongs to the neighboring halogen group.}
    \end{minipage}
\end{figure}

\begin{figure}[t]
    \begin{minipage}{0.75\textwidth}
        \includegraphics[scale=0.18]{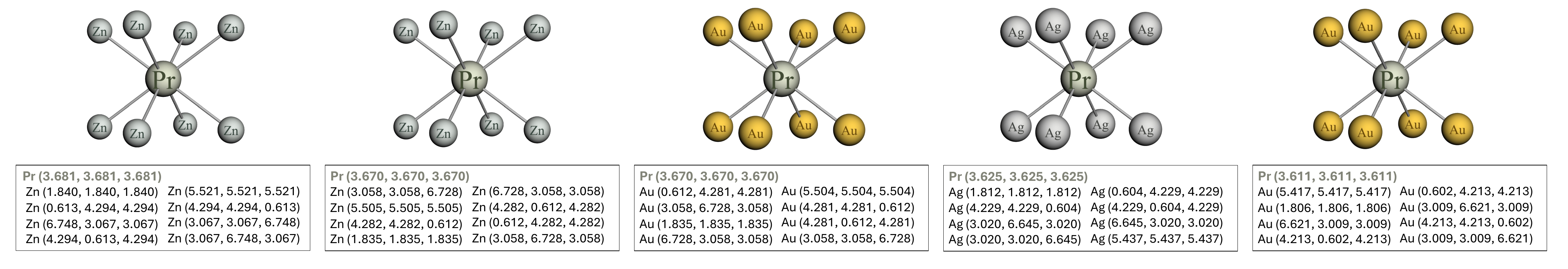}
        \text{\texttt{Au} and \texttt{Ag} are group-XI transition metals, while \texttt{Z}n belongs to the adjacent group.}
    \end{minipage}

    \vspace{1.5em}

    \begin{minipage}{0.75\textwidth}
  \includegraphics[scale=0.18]{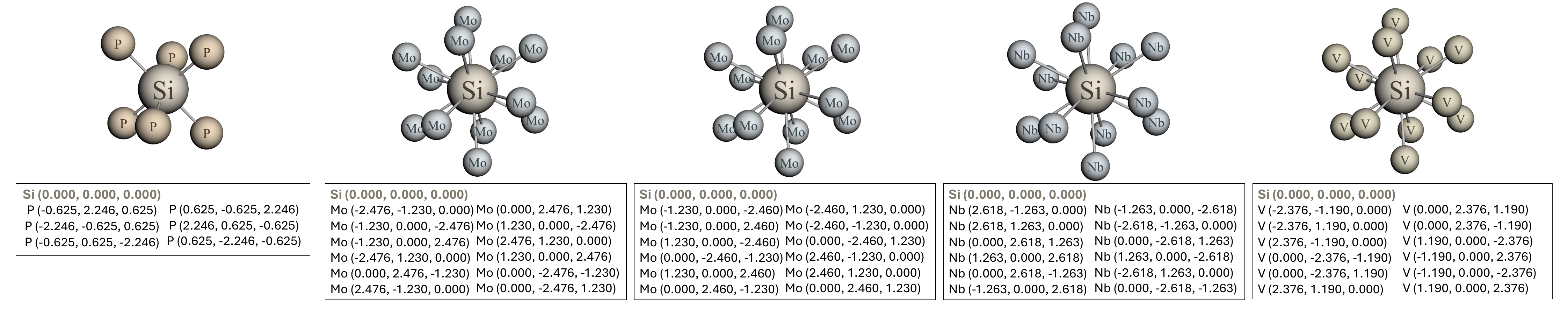}
        \text{\texttt{Nb} and \texttt{V} are group-V transition metals, while \texttt{Mo} belongs to the adjacent group.}
        
    \end{minipage}
\end{figure}
These examples show that the learned concepts encode meaningful patterns that combine both structural and elemental similarities. In particular, local environments assigned to the same concept often exhibit similar geometric arrangements, even when they originate from different chemical systems. Moreover, chemically related elements, such as elements from the same group or adjacent groups in the periodic table, frequently appear within the same concept clusters.
\section{VQ-VAE Training}
\label{vqvae_training}
\begin{figure*}[ht]
  \centering
  \includegraphics[scale=0.4]{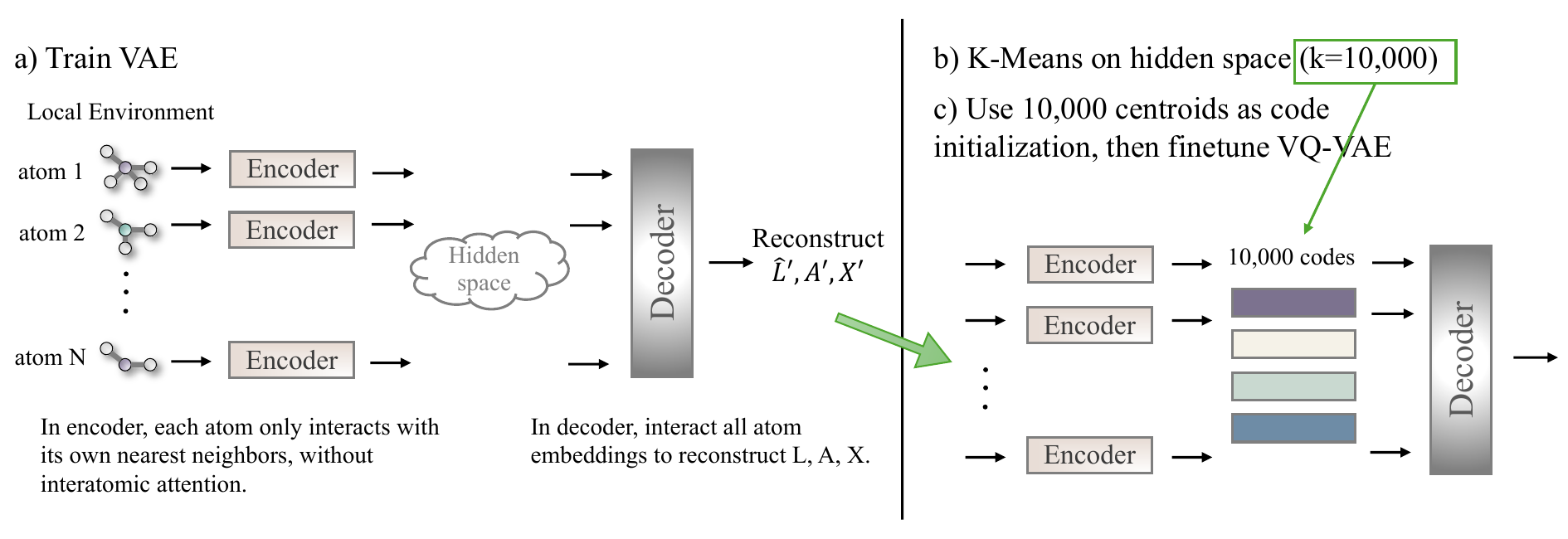}
  \caption{The whole pipeline of training VQ-VAE.}
  \label{train_vqvae}
  \vskip -0.15in
\end{figure*}
Directly training a VQ-VAE on crystal structures led to limited reconstruction performance in our preliminary experiments. Therefore, we adopt a three-stage training strategy. We first train a variational autoencoder (VAE) to learn a continuous latent space for local atomic environments. We then apply K-Means clustering to the learned latent embeddings to obtain discrete centroids. Finally, we initialize the VQ-VAE codebook with these centroids and fine-tune the full VQ-VAE model. The whole pipeline is shown in Fig.~\ref{train_vqvae}.

\paragraph{Training the VAE.}
As described in Section~\ref{Concept Extraction}, we still extract local atomic environment for each atom $i$, denoted as $\bm{T}_i\in\mathbb{R}^{(1+K)\times(3+100+6)}$. Next, we feed $\bm{T}_i$ into a VAE pipeline:
\begin{equation}
\begin{aligned}
    \label{vae}
    \{\mu_{\bm{z}_i}\},\{\log\sigma_{\bm{z}_i}\}&=\texttt{Encoder}(\{\bm{T}_i\})\in\mathbb{R}^{N\times d},\mathbb{R}^{N\times d},\\ \bm{z}_i&=\mu_{\bm{z}_i}+\sigma_{\bm{z}_i}\odot\bm{\epsilon},\bm{\epsilon}\sim\mathcal{N}(0,1)^{N\times d},\\ \bm{A}',\bm{X}',\bm{\widehat{L}}'&=\texttt{Decoder}(\{\bm{z}_i\}).
\end{aligned}
\end{equation}
Both \texttt{Encoder} and \texttt{Decoder} are implemented via the same Transformers as Fig.~\ref{transformer}(a) and (b). The training objective consists of,
\begin{equation}
\begin{aligned}
\label{vae_loss}
    &\mathcal{L}_{\bm{A}}=\frac{1}{N}\sum_i\mathrm{CrossEntropy}(\bm{A}_i,\bm{A}_i'),\quad \mathcal{L}_{\bm{X}}=\frac{1}{N}\sum_i||\bm{X}_i-\bm{X}_i'||^2_2,\\ &\mathcal{L}_{\bm{L}}=||\bm{\widehat{L}}-\bm{\widehat{L}}'||^2_2,\quad \mathcal{L}_{reg}=\texttt{KL}(\mathcal{N}(\bm{z}_i;\mu_{\bm{z}_i},\sigma_{\bm{z}_i}))||\mathcal{N}(0,1))
\end{aligned}
\end{equation}

\paragraph{K-Means Initialization.}
After training the VAE, we collect the latent embeddings of all local atomic environments from the MP-20 training set, resulting in 279,608 latent vectors. We then apply K-Means clustering using the Faiss library~\cite{faiss}, with the number of clusters set to $T=10{,}000$. The resulting cluster centroids are used to initialize the VQ-VAE codebook.

\paragraph{Fine-tuning the VQ-VAE.}
We then fine-tune a VQ-VAE by replacing the continuous latent variables of the VAE with a discrete learnable codebook. During quantization, each local atomic environment is assigned to its nearest codebook entry, and the selected code is used as a reusable structural concept. The \texttt{Encoder} and \texttt{Decoder} are initialized from the pretrained VAE, while the codebook is initialized using the K-Means centroids. The VQ-VAE pipeline and training objective are given in Section~\ref{Concept Extraction}.
\section{Experimental Details}
\subsection{Datasets Descriptions}
We conduct experiments on two datasets, MP-20 and Alex-MP-20.
\begin{itemize}
    \item \textbf{MP-20} is a standard benchmark dataset curated from the Materials Project~\cite{mp} and introduced by CDVAE~\cite{cdvae}. It contains 45,231 inorganic crystals with no more than 20 atoms per unit cell and spans 89 chemical elements. We follow the standard 60/20/20 train/validation/test split used in CDVAE. Since most MP-20 structures are experimentally known and globally stable, the dataset serves as a realistic benchmark for de novo crystal generation.

    \item \textbf{Alex-MP-20} is a larger inorganic crystal dataset introduced by MatterGen~\cite{mattergen}, constructed from the Alexandria database~\cite{alexandria} and the Materials Project~\cite{mp}. It contains 607,684 structures with at most 20 atoms per unit cell. The dataset is further filtered to retain stable or near-stable materials, typically with DFT-computed energy above hull below 0.1 eV/atom, and structures containing radioactive elements are excluded.
\end{itemize}

\subsection{Implementation Details}
\label{Implementation Details}
The encoder and decoder in the VAE/VQ-VAE are implemented as 8-layer Transformers with \#hidden\_dimension=512, \#latent\_dimension=8, \#attention\_heads=8, \#dropout=0. The maximum number of neighboring atoms is set to $\#K=12$. The regularization weights are set to $10^{-5}$ for the VAE and $10^{-2}$ for the VQ-VAE.

The composition generator $\mathcal{G}$ uses a DDPM whose denoising network is a Transformer with 12-layer=12, \#hidden\_dimension=768 and \#attention\_heads=12. The diffusion process uses \#timesteps=100 with a cosine noise schedule~\cite{Improved}.

The base generative model also adopts a DDPM, where the denoising network is an Transformer with \#layer=8, \#hidden\_dimension=512 and \#attention\_heads=8. We use diffusion \#timesteps=256 and the same cosine noise schedule~\cite{Improved}. For classifier-free guidance, the guidance strength is set to $\#\omega=2$, and the condition is randomly masked during training with probability 0.2.

\subsection{Evaluation Metric}
Our goal in de novo generation task it to generate valid, stable, unique and novel materials. These four basic metrics are defined as follows:
\begin{itemize}
    \item \texttt{Validity} (V). We consider the validity of a crystal from both structure and composition~\cite{cdvae}. For structure, a valid crystal should have volume larger than 0.1, and the minimal distance among all atom pairs should larger than 0.5. For element composition, we check charge neutrality and electronegativity difference. If one crystal satisfies these two validity simultaneously, it is overall valid.
    \item \texttt{Stability} (S). Stability measures whether a generated crystal is thermodynamically feasible. For each generated structure, we compute its energy above the convex hull, denoted as $\Delta E$, using a surrogate model (e.g., MLFF) or DFT when available. A structure is considered stable if $\Delta E<\epsilon$, where $\epsilon$ is a small threshold.
    \item \texttt{Uniqueness} (U). Uniqueness measures the diversity of generated structures by removing duplicates within the generated set. Two generated structures are considered identical if they match under the same structural equivalence criterion.
    \item \texttt{Novelty} (N). Novelty evaluates whether generated crystals are distinct from the reference dataset. A generated structure is considered non-novel if it matches any structure in the reference set under a structural equivalence criterion (e.g., using \texttt{StructureMatcher}). Otherwise, it is regarded as novel.
\end{itemize}
We discuss more on stability and novelty in this paper. To further evaluate generation quality, we further report compound metrics that measure the joint satisfaction of multiple criteria. Let $\{\mathbf{C}_{gen}\}_{i=1}^N$ denote $N$ generated crystals. For each crystal, we define indicator functions:
\begin{align}
S_i = \mathbb{I}(E_{\text{hull}}^{(i)} \leq \epsilon), \quad
N_i = \mathbb{I}(\mathbf{C}_i \notin \mathcal{D}_{\text{ref}}),  \quad
U_i = \mathbb{I}(\mathbf{C}_i \text{ is unique in } \mathcal{C}),  \quad
V_i = \mathbb{I}(\mathbf{C}_i \text{ is valid}).
\end{align}
Then, we define
\begin{itemize}
    \item \texttt{Stable \& Unique \& Novel} (S.U.N) measures the fraction of generated crystals that are simultaneously stable, unique, and novel:
    \begin{equation}
        \texttt{S.U.N} = \frac{1}{N} \sum_{i=1}^N S_i \cdot U_i \cdot N_i.
    \end{equation}

    \item \texttt{Valid \& Stable \& Unique \& Novel} (V.S.U.N) further incorporates validity, measuring the fraction of samples that satisfy all four criteria:
    \begin{equation}
        \texttt{V.S.U.N} = \frac{1}{N} \sum_{i=1}^N V_i \cdot S_i \cdot U_i \cdot N_i.
    \end{equation}
\end{itemize}

\subsection{Density Functional Theory Settings}
\label{dft}
We use DFT settings from Materials Project \url{https://docs.materialsproject.org/methodology/materials-methodology/calculation-details/gga+u-calculations/parameters-and-convergence} for structure relaxation and energy computation. In particular, we do GGA and GGA+U calculations with \texttt{atomate2.vasp.flows.mp. MPGGADoubleRelaxStaticMaker}~\citep{ganose2025_atomate2}, which in turn relies on \texttt{pymatgen.io.vasp.sets.MPRelaxSet} and \texttt{pymatgen.io.vasp.sets.MPStaticSet}~\citep{ong2013python}. Computations themselves were done with VASP~\citep{kresse1996vasp} version 5.4.4. with the plane-wave basis set~\citep{kresse1996vasp}. The electron-ion interaction is described by the projector augmented wave (PAW) pseudo-potentials~\citep{kresse1999paw}. The exchange-correlation of valence electrons is treated with the Perdew-Burke-Ernzerhof (PBE) functional within the generalized gradient approximation (GGA)~\citep{perdew1996pbegga}. The raw total energies computed by DFT were corrected with \texttt{MaterialsProject2020Compatibility} before putting into the \texttt{PhaseDiagram} to obtain the DFT $E_\text{hull}$.

We do DFT relaxation firstly using the generated crystal structures. Then, for the crystals failed to DFT, we follow previous studies~\cite{flowmm,adit} and use MatterSim-v1-1M~\cite{mattersim} to do pre-relaxation, and then redo DFT.

\subsection{Operating Environment}
\label{Operating environment}
The environment where our code runs is shown as follows:
\begin{itemize}
    \item Operating system: Linux version 6.8.0-63-generic
    \item CPU information: AMD EPYC 9554 64-Core Processor
    \item GPU information: NVIDIA Corporation AD102GL [L40S]
\end{itemize}
\section{Related Work}
\label{related work}
\paragraph{Generation on Scientific Data}
Recent progress in deep generative modeling, particularly diffusion-based methods, has greatly advanced scientific discovery tasks such as molecule and material design. AlphaFold3~\cite{abramson2024accurate} employs diffusion to achieve accurate all-atom biomolecular complex generation. GeoLDM~\cite{xu2023geometric} highlights the capability of diffusion models in scientific data by modeling 3D geometric structures and generating physically consistent molecular samples. Graph-GRPO~\cite{zhu2026graph} further enhances molecule generation by aligning graph flow models with task-specific objectives through reinforcement learning~\cite{sui2025if}.

For crystal generation, CDVAE~\cite{cdvae}, FlowMM~\cite{flowmm}, and ADiT~\cite{adit} introduce variational, flow matching, and latent diffusion frameworks, respectively. DiffCSP~\cite{diffcsp} proposes CSPNet, which has become a standard equivariant denoising backbone. Follow-up works, including DiffCSP++~\cite{diffcsppp}, SGEquiDiff~\cite{sge}, and SymmCD~\cite{symmcd}, incorporate domain-specific physical priors such as space group constraints and crystallographic symmetry. MatterGen~\cite{mattergen} further extends diffusion models to conditional generation for inverse material design guided by target properties and symmetry. In addition, recent studies explore integrating large language models with crystal generation~\cite{CrystalLLM, Flowllm, crysllmgen}.

Overall, existing approaches mainly focus on developing more sophisticated generative architectures and training objectives. In contrast, our work emphasizes improving the quality of generated samples through a screening-and-refinement pipeline. The significant gains achieved even with a simple base model suggest that such a pipeline can be broadly beneficial across different generative frameworks.

\paragraph{Transformer}
Transformers~\cite{vaswani2017attention} have emerged as a versatile architecture for modeling structured data, owing to their ability to capture long-range dependencies via self-attention. Representative developments include LLaMA~\cite{Llama}, which advances large-scale autoregressive modeling, and Vision Transformer (ViT)~\cite{vit}, which extends Transformers to image modeling and facilitates progress in multimodal learning. Transformers have also been incorporated into generative modeling, such as Diffusion Transformers (DiT)~\cite{dit}, which serve as strong backbones for diffusion models. In graph domains~\cite{zhang2025can}, Graph Transformer (GT)~\cite{gt} introduces self-attention mechanisms to graph data, while subsequent studies aim to address its limitations. For instance, CoBFormer~\cite{xing2024less} firstly reveals the over-globalization issue and enhances local inductive biases with theoretical guarantees. Specformer~\cite{bo2023specformer} makes the first attempt to learn eigenvalues interaction via Transformer, triggering the fusion of attention and spectral domain.

In this work, we employ a standard Transformer as the backbone. Investigating more advanced Transformer architectures remains an important direction for future research.


\end{document}